%% file: main.tex
\documentclass[letterpaper, 10 pt, conference]{ieeeconf}  

\usepackage{cite}
\usepackage{amsmath,amssymb}
\usepackage{graphicx,mathtools,verbatim}
\usepackage[normalem]{ulem}
\usepackage{soul}
\graphicspath{{images/}}
\input{defs}

\IEEEoverridecommandlockouts                              

\title{\LARGE \bf
Learning to Navigate Cloth using Haptics
}

\author{Alexander Clegg, Wenhao Yu, Zackory Erickson, Jie Tan, C. Karen Liu, Greg Turk
\thanks{Alexander Clegg, Wenhao Yu, Greg Turk and C. Karen Liu are with the School of Interactive Computing, Georgia Institute of Technology, Atlanta, GA., USA.}
\thanks{Zackory Erickson is with the Healthcare Robotics Lab, Georgia Institute of Technology, Atlanta, GA., USA.}
\thanks{Jie Tan is with Google Brain, Google, Mountain View, CA., USA.}
\thanks{Alexander Clegg is the corresponding author {\tt\footnotesize aclegg3@gatech.edu}.}
\thanks{This material is based upon work supported by the National Science Foundation Graduate Research Fellowship under Grant No. DGE-1650044 and NSF award IIS-1514258.}
}

\begin{document}

\maketitle
\thispagestyle{empty}
\pagestyle{empty}

\begin{abstract}


We present a controller that allows an arm-like manipulator to navigate deformable cloth garments in simulation through the use of haptic information.  The main challenge of such a controller is to avoid getting tangled in, tearing or punching through the deforming cloth.  Our controller aggregates force information from a number of \emph{haptic-sensing spheres} all along the manipulator for guidance.  Based on haptic forces, each individual sphere updates its target location, and the conflicts that arise between this set of desired positions is resolved by solving an inverse kinematic problem with constraints.  Reinforcement learning is used to train the controller for a single haptic-sensing sphere, where a training run is terminated (and thus penalized) when large forces are detected due to contact between the sphere and a simplified model of the cloth.  In simulation, we demonstrate successful navigation of a robotic arm through a variety of garments, including an isolated sleeve, a jacket, a shirt, and shorts.  Our controller out-performs two baseline controllers: one without haptics and another that was trained based on large forces between the sphere and cloth, but without early termination. 


\end{abstract}

\section{INTRODUCTION}

While research in manipulation of deformable objects has made great progress in recent years, autonomous robots today still face tremendous challenges when manipulating deformable objects in the everyday human world. Most previous work has focused on planning the trajectories of the grippers/end effectors to manipulate the object into a desired configuration. In contrast, we are interested in a different problem domain where the whole manipulator must navigate around deformable objects to achieve a geometric goal state. This problem is representative of a wide variety of robotic applications; such as a manipulator retrieving objects from foliage, a snake robot navigating through rubble, a surgical end effector moving through a patient's esophagus, or a humanoid putting on a hazmat suit.

This work focuses on one of the most challenging manipulation tasks in everyday life---dressing. The goal of the dressing task is to navigate the garment to achieve a desired relative positioning of the garment and the limb. This is a challenging task, as the motion of clothing, especially in response to contact forces, is highly complex and difficult to predict. To prevent damage to clothing and increase the chance of successful completion of the task, we posit that the haptic feedback is a vital component of any control system attempting to navigate and manipulate the state of cloth. Previous work designed specialized dressing controllers without the use of haptic perception \cite{Clegg2015}. These controllers tend to be sensitive to the initial position of the manipulator relative to the garment and the materials of the garment, even though perfect vision and augmented environment information (\eg geodesic distance encoded on the surface of the clothes) are provided.


The goal of this work is to develop a control policy capable of navigating cloth using haptic sensory input. Unlike previous work, we aim to develop a generalizable feedback policy that only leverages haptic perception and proprioception to determine the next action from an observed state. As such, our approach uses reinforcement learning to optimize a policy for a \emph{haptic-sensing sphere} as a building block. The policy makes decisions based only on the contact forces exerted on the sphere and the relative position of the sphere center to the final target without any prior knowledge or vision sensing capability. By aggregating multiple haptic-sensing spheres, the learned policy can be applied to robots with arbitrary morphologies. Based on haptic forces, each individual sphere proposes an update to its location. The conflicts that arise between this set of desired positions are resolved by solving an inverse kinematic problem with constraints. While the importance of haptics in dressing tasks seems intuitive from our own experiences, it is not clear how humans exploit haptic perception to aid in dressing. In training, we reward the manipulator for proximity to a target position and we perform early termination of a training roll-out if excessive force is sensed between the cloth and the manipulator.  This gives the learning algorithm the freedom to explore effective strategies to leverage haptics for dressing tasks.


Compounding the challenge of incorporating haptics is the computational cost of cloth simulation in a contact-rich environment, as is the case with dressing. Directly generating thousands of rollouts with cloth simulation during policy learning is computationally impractical and prone to overfitting a particular type of garment. In contrast, we hypothesize that many of the navigation tasks through deformable objects share the same fundamental skill regardless of the environments or the morphology of the robot. As such, we propose to train a sphere to move through a funnel-like geometry, which provides haptic feedback in the form of contact forces between the sphere and the funnel. Due to the simplicity of the spherical geometry, the calculation of contact force can be done analytically without the need for numerical simulation. Simplifying the task and the environment drastically accelerates the training process; but, a question then arises- how well will the policy generalize to complex environments? The primary contribution of this work is a control system capable of guiding a manipulator of arbitrary morphology to complete the dressing task using only haptic data and Cartesian task targets. We show that a simple policy trained with analytical contact forces can be directly applied to navigating physically simulated cloth. Further, though trained individually, aggregated haptic-sensing spheres can work collectively to guide the manipulator to task completion.
 

We evaluate our method by testing the policy in various dressing scenarios including an isolated sleeve, a jacket, a shirt, and shorts. In each test we vary a set of parameters, such as the initial state of the manipulator or the geometric state of the cloth, and measure the success rate of the dressing task over many trials. We also compare our method to two different baseline policies. The first baseline policy operates without the use of haptic sensing while the second one penalizes contact forces quadratically. The results show that our method has a high success rate for all tasks and outperforms both baselines by a wide margin.


\section{RELATED WORK}
Navigation applications in robotics often rely on one or more of the following assumptions: a collision-free path exits, line-of-sight sensing is available, and the environment is near-static \cite{dogar2011framework, hornung2012navigation, leeper2013arm, srinivasa2010herb, katz2014perceiving}. When navigating in a cluttered, deformable environment, such as the dressing tasks in our work, none of the above assumptions hold true.


One approach to address these issues is to incorporate haptic sensing \cite{jain2013reaching, bhattacharjee2014robotic, sygulla2016adaptive,killpack2016model}. Jain \etal \cite{jain2013reaching} showed that a manipulator can reach goal locations in a cluttered environment by using a model predictive controller (MPC) and tactile sensors over the entire arm. KillPack \etal \cite{killpack2016model} further improved the MPC by modeling the full dynamics of the robot arm instead of using a quasi-static model. These methods have the ability to predict future contact forces using a bi-linear spring contact model. In a highly deformable environment, such as the interior of an article of clothing, this simple contact model is unlikely to capture the complex contact behavior accurately. However, employing a full cloth simulation will increase the computation time significantly, rendering the MPC unable to optimize the control in real time.


The importance of haptic sensing in robotic manipulation of deformable bodies has also been recognized in the emerging area of robot-assisted dressing \cite{yamazaki2014bottom, KapustaYuBhattacharjeeLiuTurkKemp2016,GaoChangDemiris2016, yu2017haptic, erickson2017how}. While the problem of manipulating deformable cloth is relevant, our work focuses on developing controllers for \emph{self dressing tasks} \cite{ho2009character, wang2013harmonic, miguel2014towards, Clegg2015}. Clegg \etal \cite{Clegg2015} proposed a full-body self-dressing controller that exploits geodesic distance information encoded on the surface of the garment to guide motion planning. However, due to the lack of haptic sensing, their method is sensitive to cloth materials and initial conditions. Our work attempts to achieve a more robust and generalizable controller for self-dressing using haptic information. Our first attempt to include haptics as a contact force penalty term in the optimization-based inverse kinematics method proposed by \cite{Clegg2015} proved to be unsuccessful. Due to the complex deformable geometry of cloth, it is difficult to strike a balance between avoiding large contact forces and making progress toward the end-effector target. As such, we do not use manual controller design and instead propose a different approach using Reinforcement Learning.

Recent advances in Reinforcement Learning have enabled the training of complex robotic motor skills with high-dimensional continuous state and action spaces \cite{gu2016q, lillicrap2015continuous, mnih2016asynchronous, schulman2015trust, schulman2015high} and human-expert-level agents in playing Atari games \cite{dqn_nature} and Go \cite{alphago_nature}. We use Trust Region Policy Optimization with Generalized Advantages Estimation (TRPO) to train our policy \cite{schulman2015trust, schulman2015high}. TRPO has been used to learn complex motor skills such as simulated humanoid running and getting up.


Directly applying TRPO to learn navigation skills in deformable environments is impractical due to the high computational cost of simulating deformable bodies. In this work, our controller is trained to perform navigation in a static environment with simple obstacle geometry. Similar approaches that use a simplified model have been used in previous work on manipulation of deformable objects \cite{Miller2012GAR, van2010gravity, balaguer2010motion, phillips2014representation, berenson2013manipulation}. For example, Miller \etal \cite{Miller2012GAR} restricted the cloth state to be vertically hanging and used a quasi-static cloth model that neglects cloth dynamics. Phillips-Grafflin \etal \cite{phillips2014representation} used a scalar deformability field defined on the deformable object represented as voxels to approximate the penalty for deformation. These methods typically assume that the deformation encountered in the testing environment is similar to that in the training environment. In contrast, our method transfers a policy trained on a single rigid sphere moving through a static, rigid, funnel-like geometry, to a different scenario in which a manipulator with arbitrary shapes navigates through a detailed, physically simulated garment.



\section{METHODS}

The core of our method is the development of a sensory-actuator building block, which we call a ``haptic-sensing sphere'', capable of tracking a target in Cartesian space based on haptic perception and proprioception. We first describe how such a building block can be trained in a reinforcement learning framework. We then describe how these building blocks can be used to create manipulators with arbitrary morphologies and transferred to unseen deformable environments. During policy execution, each haptic-sensing sphere will individually suggest an action that avoids tearing the cloth while moving towards its goal. We then apply inverse kinematics (IK) optimization to find an optimal joint configuration for the manipulator that best coordinates these independent movement suggestions. The following sections provide additional details on each of the previously mentioned system components.


 
\subsection{Training a haptic-proprioception policy} 
\label{sec:training}

\begin{figure}[t]
\vspace{3mm}
\centering
\includegraphics[width=0.45\textwidth]{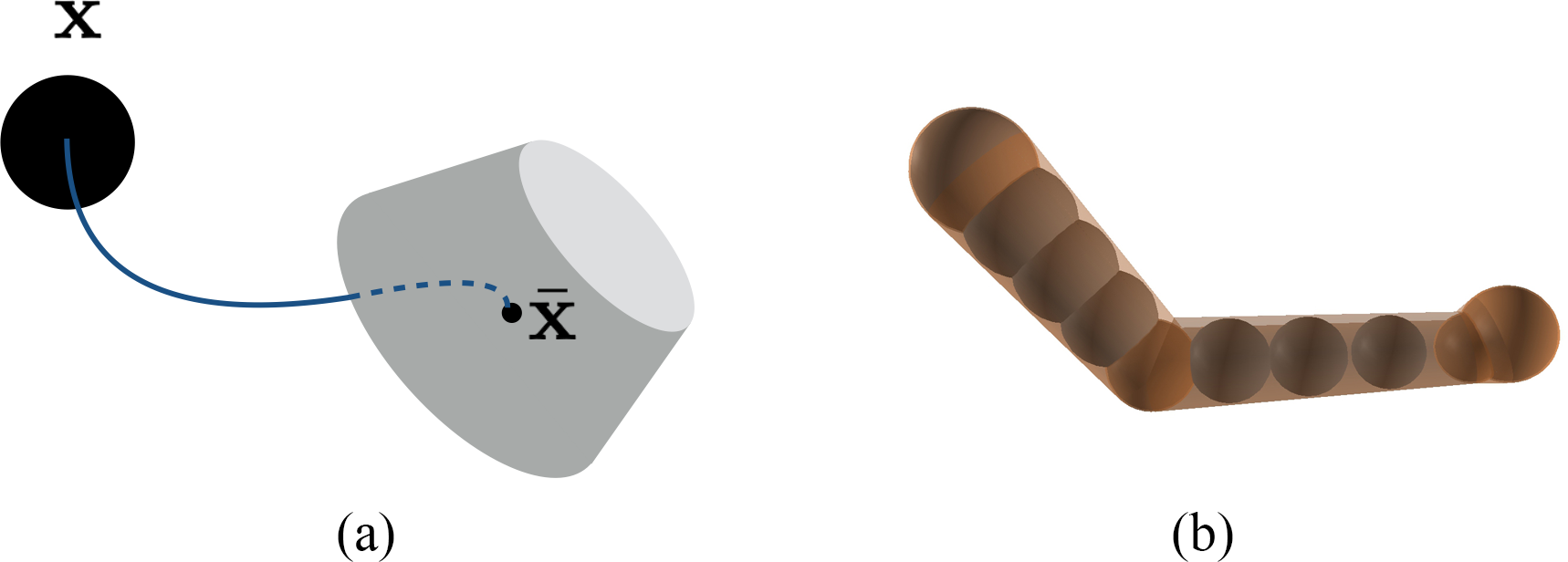}
\caption{(a) A haptic feedback controller must move a sphere with radius r, located and $\mathbf{x}$ through a stationary rigid funnel to reach the target $\mathbf{\bar{x}}$. (b) Rendering of an arm model populated with haptic feedback controller spheres. Sphere radius is for visualization purposes only and does not affect control.} 
\label{fig:funnel_arm}
\end{figure}

We observe that most dressing tasks have two goals: 1) achieve a desired relative positioning of the garment and the limb and 2) avoid excessive contact forces that could tear the cloth. To learn the fundamental skill necessary to achieve both goals, we train a haptic-sensing sphere to reach a target location at the center of a stationary rigid funnel using only the contact force and the relative position from the target (Fig.~\ref{fig:funnel_arm}). As the haptic-sensing sphere moves, it may come in contact with the funnel. If the depth of penetration between the sphere and the funnel exceeds a predefined threshold, indicating an unsafe amount of contact force has been exerted, the task is deemed unsuccessful. The task is successful if the haptic-sensing sphere reaches the target location within a certain amount of time.


To learn a control policy for the haptic-sensing sphere, we formulate a Markov Decision Process (MDP) defined by a tuple $(S, A, P, R, \gamma, \rho)$, where $S$ is the \emph{state} space, $A$ is the \emph{action} space, $P: S\times A \times S\mapsto \mathbb{R}$ is the \emph{transition function}, $R$ is the \emph{reward function}, $\gamma \in [0, 1]$ is the \emph{discount factor}, and $\rho$ is the \emph{distribution of the initial states}. MDP solves for a stochastic \emph{policy} $\pi : S \times A \mapsto \mathbb{R}$ that maximizes the \emph{expected return}. 
\begin{displaymath}
\mathop{\mathbb{E}}_{s_0,a_0,...}\left[\sum_{t=0}^{\infty}\gamma^tR(s_t)\right]
\end{displaymath}
where $s_0\sim\rho(s_0), a_t\sim\pi(a_t|s_t), s_{t+1}\sim P(s_{t+1}|s_t, a_t)$.

In our problem, a state, $[\mathbf{\bar{x}}-\mathbf{x}, \tilde{\vc{f}}]$ includes the relative position from the current center of the haptic-sensing sphere $\mathbf{x}$ to the target $\mathbf{\bar{x}}$, and a normalized contact force computed analytically $\tilde{\vc{f}} = \frac{d}{r}\mathbf{n}$, where $r$ is the radius of the sphere, $d$ is the penetration depth between the sphere and the funnel and $\mathbf{n}$ is the direction of penetration. The action is simply the velocity $\mathbf{v}$ that moves the haptic-sensing sphere to its next location. In our case, the transition function $P$ is deterministic. We numerically integrate the action $\mathbf{v}$ to calculate the new position of the sphere $\vc{x}^{next} = \vc{x} + \Delta t \vc{v}$, from which the new state can be computed.



We use a neural network to represent the policy $\pi$. The roll-out terminates immediately if the haptic-sensing sphere successfully reaches the target $||\mathbf{x}-\mathbf{\bar{x}}||_2<\epsilon$ or the penetration is too deep $d>0.5r$. At each state, the reward function $R$ penalizes the distance to the goal and the failure of the task:
\begin{displaymath}
R(s)=-||\mathbf{x}-\mathbf{\bar{x}}||_2+\left\{
\begin{array}{ll}
5 & \textrm{if sphere at target,}\\
-10 & \textrm{if penetration too deep,}\\
0 & \textrm{otherwise.}
\end{array}
\right.
\end{displaymath}

The neural network is trained using Trust Region Policy Optimization (TRPO) \cite{schulman2015trust} with a curriculum learning strategy \cite{Bengio:2009}. We start with a shallow funnel, with which TRPO can find a successful policy from random initialization. We then continue training with a wider funnel, as shown in Fig. \ref{fig:funnel_arm}a, which improves the robustness of the learned haptic feedback controller. During training with a funnel, both the initial position of the haptic-sensing sphere and the orientation of the funnel are chosen at random. This helps to train a controller that can handle a variety of goal directions and obstacle orientations.


\subsection{Controlling a Full Limb}

Once trained, multiple haptic-sensing spheres can be aggregated to control a manipulator in a dressing scenario. We represent a manipulator as connected capsules (Fig. \ref{fig:funnel_arm}b) and place multiple haptic-sensing spheres along the medial axis of each capsule. Given the contact forces between the manipulator and the cloth, each haptic-sensing sphere queries its own policy based on the current state, $[\mathbf{\bar{x}}_i - \mathbf{x}_i, \tilde{\vc{f}}_i]$, to provide an independent suggestion on how to move the manipulator during dressing.


To compute the contact force, $\tilde{\vc{f}}_i$, exerted on the $i$-th haptic-sensing sphere, we use PhysX cloth simulator \cite{Macklin:2014} to provide collision detection and resolution between the manipulator and every vertex on the cloth. These per-vertex contact forces are aggregated to compute the contact force on the $i$-th haptic-sensing sphere:
\begin{equation}
\tilde{\vc{f}}_i=\frac{\sum_{j\in\Omega_i} \mathbf{f}_j}{F_{max}}
\label{eq:forceRatio}
\end{equation}
where $\mathbf{f}_j$ is the contact force at the vertex $\vc{v}_j$. $\Omega_i$ is a set of vertex indices satisfying:
\begin{equation}
j \in \Omega_i, \;\;\mathrm{if} \;i = \argmin_k d(\vc{x}_k, \vc{v}_j)
\end{equation}
where $d$ is a function that computes the Euclidean distance between the center of the $k$-th sphere and the $j$-th vertex. This operation results in contact forces from collision between the character geometry and the cloth being binned into the nearest haptic sensor. Note that while changes in contact geometry and the placement of haptic-sensing sphere centers will affect this process, the radius of the haptic-sensing spheres will not.

$F_{max}$ is the maximum contact force that can be exerted on the cloth without tearing it. This value can be determined by the user or measured empirically from the cloth simulator and accounts for variation in garment mass, size and material properties. With the normalization of $F_{max}$ in Equation \ref{eq:forceRatio}, the magnitude of contact force exerted on a haptic-sensing sphere always ranges from 0 to 1.

In addition to the normalized contact force, we also need a target $\mathbf{\bar{x}}$ as the input for each individual haptic controller. For this purpose, we classify the haptic-sensing spheres into two types: \emph{leading} controllers and \emph{trailing} controllers. In our examples, the controller that is located at the end effector is the leading controller and the rest are trailing controllers. The target of the leading controller is task specific and given by the user. For example, if the task is to stretch the manipulator through a sleeve, the target position would be at the cuff. The target positions of the trailing controllers are their current location $\mathbf{\bar{x}}=\mathbf{x}$. In other words, the trailing haptic controllers will try to stay stationary while avoiding exerting too much force on the cloth. With the current state $[\mathbf{\bar{x}}_i - \mathbf{x}_i, \tilde{\vc{f}}_i]$ as the input, the policy of each haptic-sensing sphere will compute an action, $\mathbf{v}_i$, which is the displacement between the current and next locations: $\vc{x}_i^{next} = \vc{x}_i + \Delta t \vc{v}_i$.

Since the action for each sphere is computed without respecting the kinematic constraints of the manipulator, we need to reconcile the suggested motions from the individual spheres. We use an inverse kinematics (IK) solver to find an optimal joint configuration $\mathbf{q}^*$ which best match the collective output of all the haptic-sensing spheres:
\begin{equation}
\mathbf{q}^*=\argmin_\vc{q} {\sum_i{w_i||\mathbf{p}(\mathbf{q},\mathbf{r}_i)-(\mathbf{x}_i+\Delta t\mathbf{v}_i)||^2}}
\label{eq:IK}
\end{equation}
where $\mathbf{r}_i$ is the local coordinate of $i$-th haptic-sensing sphere on the manipulator, which is transformed to the world coordinate by $\mathbf{p(q,r_i)}$, and $\mathbf{x}_i+\Delta t\mathbf{v}_i$ is the new location suggested by the $i$-th haptic-sensing sphere. The weight $w_i$ specifies the relative importance of each sphere. In all our experiments, we set the weight of the leading controller to be 40 times larger than those of the trailing controllers. The optimization (\ref{eq:IK}) is solved by gradient descent. Once the desired configuration $\vc{q}^*$ is solved, we kinematically adjust the manipulator to $\vc{q}^*$ and continue to simulate the cloth.

\section{RESULTS}

The motions of the manipulators in this work are simulated by Dart\cite{DART}, which is a multi-body physics simulator supported by Gazebo. The cloth is simulated using an implementation of position-based dynamics via PhysX\cite{Macklin:2014} and the garments are represented as triangle meshes with a default cloth material. The haptic feedback control policy is represented by a Multilayer Perceptron neural network with two hidden layers, each consists of $32$ hidden units with tanh activation functions. The learning process takes $500$ iterations of TRPO updates. During each iteration, $4000$ steps are simulated. We limit rollout length to $1000$ steps for each sample. In order to train a control policy that is invariant to the target position direction and the force direction, we randomly sample the orientation of the training funnel geometry and uniformly initialize the sphere in a $1m\times1m\times1m$ box centered at the origin. We train the policy for a sphere of radius $0.2m$ and funnel approximately twice that radius. 

To evaluate the effectiveness of our proposed approach, we examine four representative dressing scenarios with increasing difficulty; namely a sphere traveling linearly through a cloth tube, dressing a jacket, dressing a pair of shorts and dressing a T-shirt. The goal of this evaluation is to show that given minimal task specific input, our haptics-informed control architecture enables arbitrary limb morphologies to robustly navigate a variety of garments without exerting excessive force. In each case, a guiding path is provided that the manipulator tracks (shown as purple curves in Figures \ref{fig:tube} to \ref{fig:tshirt_arm}). These paths are interpolating cartesian splines formed by connecting the centroids of user-defined vertex loops on the garment. For example, one control point may be the centroid of all vertices forming the end of one sleeve. As these vertices move during simulation, the control curve is re-computed and the task target is updated. To assess the robustness of our controller in performing the dressing tasks, we add variations to the initial conditions for each trial and average the performance of all sampled rollouts to obtain the final success rate. Each simulation rollout's control cycle is preceded by 2 seconds of cloth simulation to settle the garment followed 2 seconds of joint pose interpolation during which the limb is moved into its randomly drawn initial condition. Below are details about the four dressing tasks, including variations of initial conditions. We encourage the reader to view examples of these evaluations and additional tasks in our supplemental video.

\begin{table}[t!]
\vspace{3mm}
\caption{Performance comparison on the four dressing tasks with a friction value of 0.2.}
\vspace{-3mm}
\label{table_example}
\begin{center}
\begin{tabular}{|c|c|c|c|c|c|c|}
\hline
 & \multicolumn{2}{c|}{Proposed} & \multicolumn{2}{c|}{Baseline 1} & \multicolumn{2}{c|}{Baseline 2}\\
\hline
 & SR & TC & SR & TC & SR & TC \\
\hline
Cloth Tube & 100\%& 4.43s& 100\%& 4.32s& 32\%& 5.52s\\
\hline
Jacket & 100\%& 6.30s&17\% &6.42s &86\% & 6.55s\\
\hline
Shorts & 96\% & 6.52s& 0\% & N/A & 91\% & 6.49s\\
\hline
T-shirt & 92\% & 8.87s & 0\%& N/A & 0\% &  N/A\\
\hline

\end{tabular}
\end{center}
\label{tbl:performance}
\end{table}

\begin{itemize}

\begin{figure*}[t!]
\centering
\includegraphics[width=\linewidth]{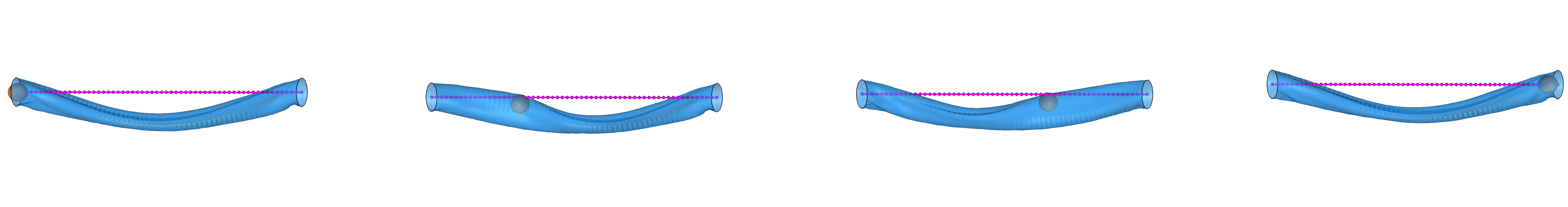}
\caption{Result of cloth tube traversal with our controller.}
\label{fig:tube}
\vspace{-3mm}
\end{figure*}

\begin{figure*}[t!]
\centering
\includegraphics[width=\linewidth]{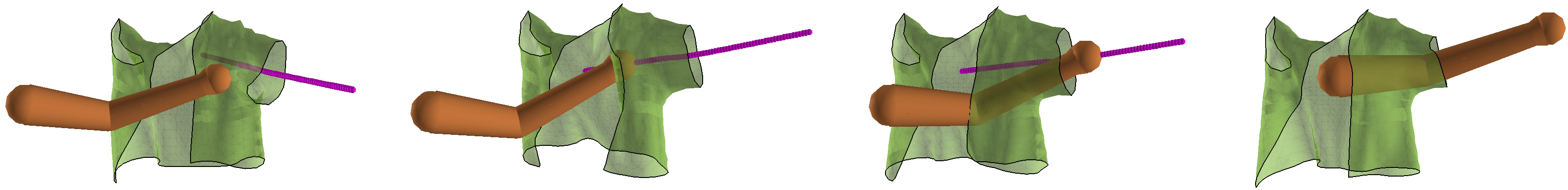}
\caption{Result of jacket dressing with our controller.}
\label{fig:jacket}
\vspace{-3mm}
\end{figure*}

\begin{figure*}[t!]
\centering
\includegraphics[width=\linewidth]{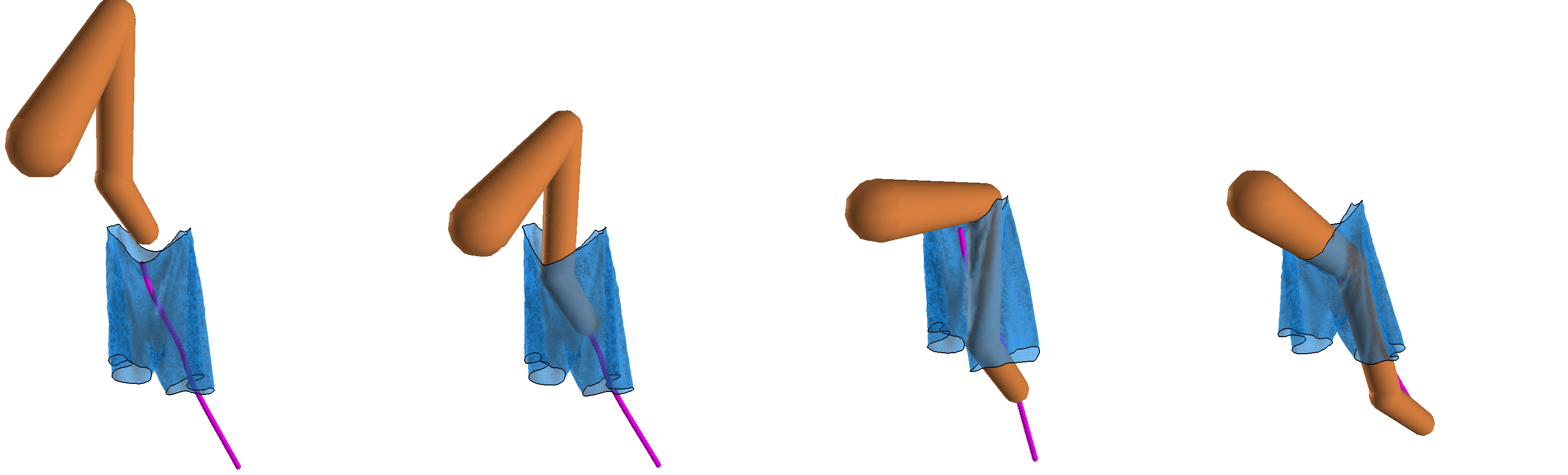}
\caption{Result of shorts dressing with our controller.}
\label{fig:shorts}
\vspace{-3mm}
\end{figure*}

\begin{figure*}[t!]
\centering
\includegraphics[width=\linewidth]{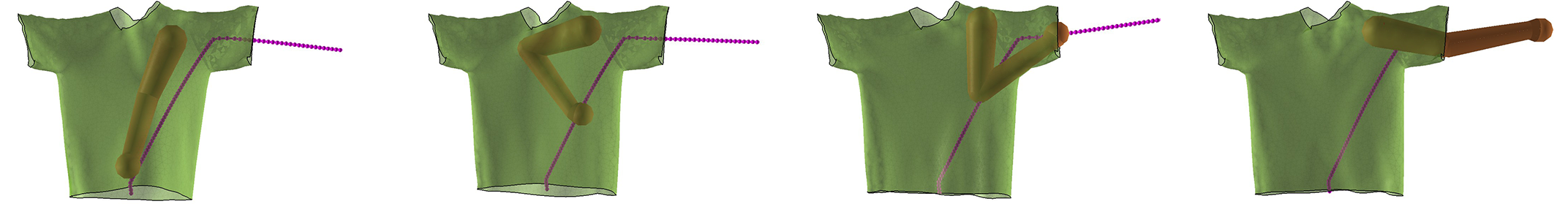}
\caption{Result of T-shirt dressing with our controller.}
\label{fig:tshirt_arm}
\vspace{-3mm}
\end{figure*}

\begin{figure*}[t!]
\centering
\includegraphics[width=\linewidth]{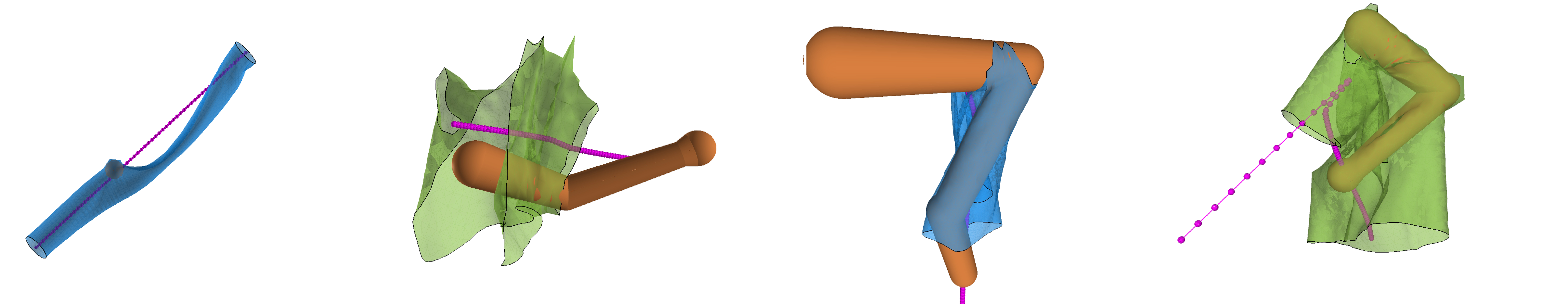}
\caption{Examples of Baseline 1 failing to complete dressing tasks.}
\label{fig:fail}
\vspace{-3mm}
\end{figure*}

\item{\textbf{Cloth tube:}  A sphere begins at one end of a cloth tube and follows a fixed linear target trajectory to the other end. The orientation of the tube is chosen from a uniform spherical distribution and varies relative to the gravity direction, resulting in a range of drape configurations. There is no variation in initial sphere position in this example. The fixed linear target trajectory through the draped regions guarantees that the sphere must push the garment out of the way in order to reach the task goal (Fig. \ref{fig:tube})}
\item{\textbf{Jacket:} An arm with approximately human joints and proportions enters through the front of a jacket, and navigates into and out of one sleeve. The initial position and joint angles of the arm are varied roughly 17 degrees around a rest pose and it occasionally starts in contact with the garment. The arm must move around and push through the hanging jacket body to reach the sleeve. This task is fairly easy, as the arm is relatively unconstrained with a large translational range at the shoulder and the jacket sleeve acts similarly to a loosely hanging cloth tube (Fig. \ref{fig:jacket})}
\item{\textbf{Shorts:} A leg with a translation limited hip and approximately human joints and proportions is initialized above a pair of shorts. The toe is guided to pass into one leg of the shorts. There is variation in the hip translation and roughly 6 degrees of variation in all initial joint degrees of freedom of the leg. In this example, the foot must pass through the waistband without catching it with the heel. The leg must also extend in response to forces on the lower thigh and shin in order to respect the in-extensibility of the pinned garment. However, the shorts hang open in the gravity direction, making this a medium difficulty example (Fig. \ref{fig:shorts})}
\item{\textbf{T-shirt:} The same arm as in the jacket task, now with more limited translation at the shoulder is initialized inside a T-shirt. The arm is guided to pass up through the body, into and out of the shirt sleeve nearest the pinned shoulder. The initial joint angles of the arm and translation range of motion of the shoulder are varied within roughly 17 degrees, as well as a larger variation in the elbow angle. In order to complete this task, the arm must first directly oppose the end effector motion, pushing the elbow against the back of the shirt, and later must pivot the upper arm around sleeve contact points on the forearm in order to push the arm out of the sleeve. The highly constrained nature of the workspace makes this a challenging task (Fig. \ref{fig:tshirt_arm})}

\end{itemize}

We compare our controller with two baselines: a haptic-unaware controller (Baseline 1) which moves linearly toward the target at a fixed speed and a haptic feedback controller trained with an additional penalty on the magnitude of force in the reward function (Baseline 2). We compare two metrics across the three controllers: success rate (SR), which is computed as the number of successful trials divided by the total amount of trials, and the time to completion (TC), which is the average time to complete the dressing task for the successful trials. All trials are given 10 simulator seconds to complete the navigation task after which the simulation is terminated and counted as failure. The results are shown in Table \ref{tbl:performance} and each table entry is the result of 100 trials. 


From Table \ref{tbl:performance}, we can see that our approach outperforms the baseline controllers in all four dressing tasks. The T-shirt dressing task, which involves navigating the end-effector through the garment while having the upper arm inside the garment, poses significant difficulties for the baseline controllers, resulting in zero successful trials. In contrast, the controller trained using our method still achieves high performance with a success rate of $92\%$. Examples of situations when the baseline controllers fail to perform the dressing tasks are shown in Fig. \ref{fig:fail}.


 
\begin{figure}[t!]
\centering
\includegraphics[width=\linewidth]{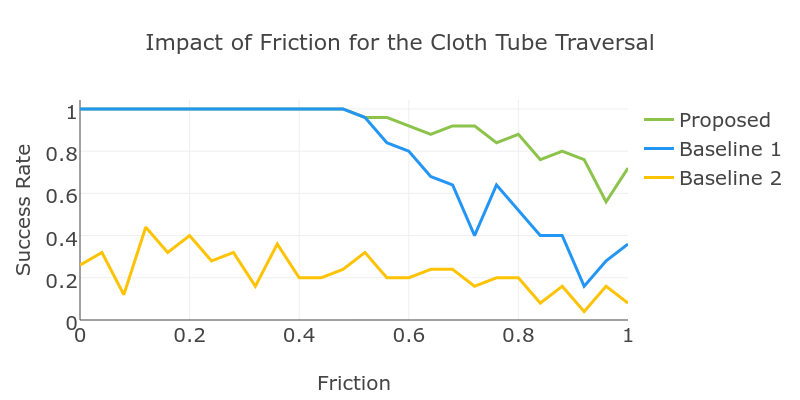}
\caption{Impact of friction for the cloth tube traversal. For each controller, we uniformly sampled 26 friction coefficients and averaged the success rate for each coefficient over 25 trials.}
\label{fig:sphere_friction}
\vspace{3mm}
\end{figure}

To examine how our algorithm generalizes to different cloth materials, we perform the cloth tube traversal task with different friction coefficients between the end-effector and the garment. The success rate of the three different controllers can be seen in Fig.\ref{fig:sphere_friction}.  In the low friction case, both Baseline 1 and our controller perform well.  As friction increases, it becomes more difficult to navigate the cloth tube, and the success rates become lower. However, there remains a key difference in the performance of our proposed controller and the haptic unaware baseline 1. When baseline 1 fails, it is due to garment tearing, whereas our controller refuses to continue moving forward and stalls until the time limit is reached. This is an important feature, as any cloth navigation controller's first priority should be to respect force limits.

Although baseline 2 only completes the tube navigation task at the easiest cloth orientations (nearly vertical), it succeeded at never tearing the cloth. This result is consistent with the expected training outcome of a reward function that penalizes all force and therefore results in a controller that is unwilling to push on the cloth even when the forces are low. While this policy is quite successful during analytical funnel training, it fails to generalize to cloth navigation. This also explains the slightly poorer performance of this baseline on the jacket example, where the limb must push through the sleeve opening.  This baseline performs better on the shorts example, where the garment more closely resembles a funnel and forces are only needed to avoid tearing the garment. We refer the reader to our supplemental video for examples of the discussed results. 
   
\section{CONCLUSION}

We have presented a haptic controller that allows a manipulator to navigate through a variety of deformable cloth geometries, including shirt, jacket and pants.  A key aspect of our approach is that by assembling multiple haptic-sensing spheres, each part of the manipulator can detect and react to collisions with the cloth.  Due to the modular nature of our manipulators, we can create various manipulator shapes: a single sphere, an arm, a leg, and an upper torso (head together with two arms).  For all of these examples, our controller out-performs two baseline controllers in terms of frequency of task completion.

Despite the success of our controller, there are several avenues for future work.  First, our controller is only capable of responding to a deformable environment, and is not capable of high-level planning.  A logical extension of our work would be to incorporate our controller into a system that can make higher level plans, such as making decisions about which direction to tuck an elbow or when to backtrack when the end effector is stuck.  A second limitation of our approach is that we do not yet take into account the possibility of self-collisions and kinematic constraints other than joint limits and rigid connections. For most applications, it will be necessary to resolve potential collisions between the manipulator and the robot or an animated human body.  Finally, we would like to deploy our controller on a real robot that interacts with cloth. However, there are several challenges that face implementation on a physical robot. First, this will require the development of haptic sensors that are small enough to be densely distributed along a manipulator. Additionally, these sensors must be sensitive enough to detect the small forces that occur when interacting with clothing.







\section*{ACKNOWLEDGMENTS}

We thank Charlie Kemp for his feedback on this work.


\bibliographystyle{IEEEtran}
\bibliography{reference_bib}

\end{document}

%% file: defs.tex
\usepackage[dvipsnames]{xcolor}

\newcommand{\zackory}[1]{\textcolor{Orange}{{Zackory: #1}}}

\long\def\ignorethis#1{}

\newcommand{\etal}{{\em{et~al.}\ }}
\newcommand{\eg}{e.g.\ }

\newcommand{\vc}[1]{\ensuremath{\mathbf{#1}}}

\newcommand{\argmin}{\operatornamewithlimits{argmin}}

\newcommand{\pctab}{\hspace{0.2in}}

